\pdfoutput=1

\documentclass[11pt]{article}

\usepackage{ACL2023}
\usepackage{url}
\usepackage{multirow}
\usepackage{booktabs}
\usepackage{bbding}

\usepackage{times}
\usepackage{latexsym}

\usepackage{pgfplots}
\usepackage{graphicx}
\usepackage{subfigure}

\usepackage{algorithm}
\usepackage{algorithmic}

\usepackage[T1]{fontenc}

\usepackage[utf8]{inputenc}

\usepackage{microtype}

\usepackage{inconsolata}

\title{SAMP: A Model Inference Toolkit of Post-Training Quantization for Text Processing via Self-Adaptive Mixed-Precision}

\author[1 \thanks{\ \ Corresponding author: Rong Tian. \\ \textnormal{E-mail: tianrong03@kuaishou.com}}]{Rong Tian}
\author[2]{ Zijing Zhao}
\author[2,3]{Weijie Liu}
\author[3,4]{Haoyan Liu}
\author[3]{\\Weiquan Mao}
\author[3]{Zhe Zhao}
\author[1]{Kan Zhou}

\affil[1]{Kuaishou Technology, Beijing, China}
\affil[2]{Peking University, Beijing, China}
\affil[3]{Tencent AI Lab, Beijing, China}
\affil[4]{Institute of Dataspace, Hefei Comprehensive National Science Center, Anhui, China}

\begin{document}
\maketitle
\begin{abstract}
The latest industrial inference engines, such as FasterTransformer\footnote{\url{https://github.com/NVIDIA/FasterTransformer}\label{footnote:FT}} and TurboTransformers \citep{fang2021turbotransformers}, have verified that half-precision floating point (FP16) and 8-bit integer (INT8) quantization can greatly improve model inference speed. However, the existing INT8 quantization methods are too complicated, and improper usage will lead to model performance damage greatly. In this paper, we develop a toolkit for users to easily quantize their models for inference, in which \textbf{S}elf-\textbf{A}daptive \textbf{M}ixed-\textbf{P}recision (SAMP) is proposed to automatically control quantization rate by a mixed-precision architecture to balance model accuracy and efficiency. Experimental results show that our SAMP toolkit has a higher speedup than PyTorch \citep{paszke2019pytorch} and FasterTransformer while ensuring the required accuracy. In addition, SAMP is based on a modular design, decoupling the tokenizer, embedding, encoder and target layers, which allows users to handle various downstream tasks and can be seamlessly integrated into PyTorch.
\end{abstract}


\section{Introduction}
\begin{table*}[]
\resizebox{1.0\textwidth}{!}{
\begin{tabular}{@{}l|c|ccc|ccc@{}}
\toprule
\multirow{2}{*}{\textbf{Inference Toolkit}} & \multirow{2}{*}{\textbf{Tokenizer}} & \multicolumn{3}{c|}{\textbf{Mixed-Precision GEMMs}} & \multicolumn{3}{c}{\textbf{Downstream Tasks}}                        \\ 
                                     & & Layers  & MHA-FFN  & Fully-quantized & Classification & NER & Text Matching   \\ \cmidrule(r){1-8}
FasterTransformer                    & \XSolidBrush              & \XSolidBrush     & \XSolidBrush     & \CheckmarkBold           & \CheckmarkBold              & \XSolidBrush          & \XSolidBrush          \\
TurboTransformers                    & \XSolidBrush              & \XSolidBrush     & \XSolidBrush     & \XSolidBrush          & \CheckmarkBold              & \XSolidBrush          & \XSolidBrush           \\
LightSeq                             & \XSolidBrush              & \XSolidBrush     & \XSolidBrush     & \XSolidBrush         & \XSolidBrush              & \XSolidBrush          & \XSolidBrush            \\
SAMP                         & \CheckmarkBold              & \CheckmarkBold     & \CheckmarkBold     & \CheckmarkBold        & \CheckmarkBold              & \CheckmarkBold          & \CheckmarkBold            \\ \bottomrule
\end{tabular}
}
\caption{
Features for FasterTransformer, TurboTransformers, LightSeq and our proposed SAMP. SAMP supports tokenizer, different combinations of mixed-precision modes and various downstream tasks.
}
\label{sec:toolkit_table}
\end{table*}

Text understanding is one of the basic tasks in the field of Natural Language Processing (NLP), including information retrieval, dialogue system, sentiment recognition, summarization, language model, etc. Transformer-based models \citep{vaswani2017attention} have achieved state-of-the-art in many downstream tasks, such as BERT \citep{devlin2018bert}, XLNet \citep{yang2019xlnet}, Google T5 \citep{raffel2020exploring}, etc. In some large industrial systems, training frameworks (e.g. TensorFlow \citep{abadi2016tensorflow} or PyTorch \citep{paszke2019pytorch}) are not good options to deploy models due to the lack of high GPU occupation considerations and good memory management of them during the inference phase \citep{wang2021lightseq}.



Conventional inference acceleration tools for deep learning models such as NVIDIA TensorRT \citep{vanholder2016efficient}, TurboTransformers \citep{fang2021turbotransformers} and LightSeq \citep{wang2021lightseq} are primarily designed for fixed-size or variable-length inputs.
These tools' optimization concepts mainly take into account memory management, operation fusion, or other data pruning techniques in the online computing systems, mostly single-precision calculation (only floating-point is used). So the acceleration performance is limited. On this basis, FasterTransformer developed by NVIDIA performs fixed-point acceleration on Transformer models (using Fully-quantization in all transformer layers), and has achieved an excellent speedup compared with floating-point. However, this method of Fully-quantized in all transformer layers makes it difficult for INT8-quantization inference results to achieve the accuracy of floating-point calculations, resulting in a large loss of calculation accuracy in specific tasks, making it difficult to be widely used. On the other hand, we find that the kernel-fusion policy in FasterTransformer INT8-quantization implementation can still be optimized.

To solve these problems, we propose an inference toolkit SAMP, which contains a self-adaptive mixed-precision Encoder and a series of advanced fusion strategies. Objectively, The mixed-precision calculation of floating-point and fixed-point can obtain better calculation accuracy than fully-fixed-point calculation.
\textbf{Self-Adaptive Mixed-Precision Encoder} can find an optimal combination of mixed-precision among a large number of General Matrix Multiplication (GEMM) operations and Transformer layers, which can align the performance of model inference most closely with user needs (calculation accuracy and inference latency).
\textbf{Advanced Fusion Strategies} make fusion improvements for embedding kernels and quantization-related operations respectively, reducing CUDA kernel calls by half. Moreover, SAMP is an end-to-end toolkit implemented by C++ programming language (from Tokenizer to Embedding, Encoder, and Downstream tasks), which has excellent inference speed and reduces the threshold of industrial application. Table~\ref{sec:toolkit_table} shows the innovative features compared with similar systems. We present the following as the key contributions of our work:

\begin{description}
  \item[Self-Adaptive] SAMP balances computational accuracy and latency performance in post-training quantization inference methods.
  Users can choose a mixed-precision configuration with appropriate accuracy and inference latency for different tasks. SAMP also suggests a combination of quantization modes automatically via an adaptive allocation method.
  
  \item[Efficiency] SAMP shows better inference speedup than other inference toolkits in a wide precision range (from floating-point to fixed-point). In CLUE\footnote{\url{https://github.com/CLUEbenchmark/CLUE}\label{footnote:CLUE}} classification task datasets, SAMP achieves up to 1.05-1.15 times speedup compared with FasterTransformer.

  \item[Flexibility] SAMP covers lots of downstream tasks, such as classification, sequence labeling and text matching. And Target modules are extensible and flexible to customize.
  It is user-friendly and less dependent. SAMP supports both C++ and Python APIs, only requires CUDA 11.0 or above. We also provides many convenient tools for model conversion.
\end{description}


\section{Related Work}


\subsection{Quantization in Neural Networks}
Quantizing neural networks dates back to the 1990s \citep{balzer1991weight, marchesi1993fast}. In the early days, the main reason to quantize models is to make it easier for digital hardware implementation \citep{tang1993multilayer}. Recently, the research of quantizing neural networks has revived due to the success of deep learning \citep{guo2018survey}. A slew of new quantization methods have been proposed, which are divided into two categories, post-training quantization (PTQ) and quantization-aware training (QAT), according to whether the quantization procedure is related to model training. PTQ requires no re-training and is thus a lightweight push-button approach to quantization, only calibration needed. 
QAT requires fine-tuning and access to labeled training data but enables lower bit quantization with competitive results \citep{jacob2018quantization}. However, this method is difficult to popularize in the industry, especially for many existing models that need to be retrained, and the training process is also very long. Both FasterTransformer and our SAMP use PTQ to achieve fixed-point quantization acceleration.

\subsection{Kernel Fusion}
Kernel fusion can improve computational efficiency by reducing the number of memory accesses, increasing cache locality and reducing kernel launch overhead \citep{fang2021turbotransformers}.
Especially in inference, because of no back-propagation procedure, some small adjacent operators can be fused into a larger kernel.
Previous fusion methods mainly include Tensor-fusion and Layer-fusion \citep{vanholder2016efficient}. Tensor-fusion is mainly to concatenate tensors of the same shape into one tensor. Layer-fusion is to fuse operators of adjacent layers into one operator layer. 
Our fusion improvement of embedding kernels in SAMP is Tensor-fusion, and the operation fusion of quantization operators belongs to Layer-fusion.

\section{Architecture}
\subsection{Overview of SAMP}
\begin{figure}[htbp]
    \centering
    \includegraphics[width=0.45\textwidth]{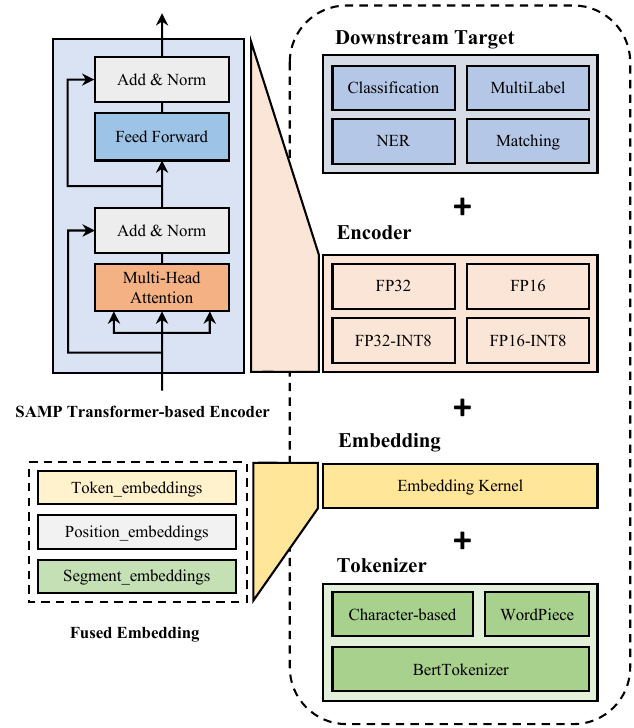}
    \caption{The architecture of SAMP.}
    \label{fig:samp_arch}
\end{figure}

In this section, we mainly introduce four modules of SAMP as shown in Figure~\ref{fig:samp_arch}: Tokenizer, Embedding, Encoder and Downstream Target, and some innovative features make it stand out from other similar works.

\textbf{Tokenizer:}
SAMP is a task-oriented and end-to-end inference library, which has a complete word segmentation module for Chinese and English that supports multi-granularity tokenization, such as character-based tokenization, wordpiece tokenization and general BertTokenizer. This module is implemented by C++ programming language and multi-thread processing methodology. So, its processing is faster than some Python programming language implementations in similar inference libraries.

\textbf{Embedding:}
Current embedding method proposed by BERT \citep{devlin2018bert} is constructed by summing the corresponding token, segment, and position embeddings, which is implemented by previous work (e.g. FasterTransformer) through three independent operation kernels. We fuse these three operators into one kernel (Embedding Kernel) to reduce CUDA kernel calls, as shown in Figure~\ref{fig:samp_arch}.

\textbf{Encoder:}
SAMP selects Transformer \citep{vaswani2017attention} as the basic component in Encoder module. Our innovative features about how to quantitatively balance the accuracy and latency performance of FP16 and INT8 are mainly realized in this module, and we propose an \textbf{Accuracy-Decay-Aware allocation} algorithm to obtain best speedup of mixed-precision while ensuring the required accuracy automatically. At the same time, the fusion improvements of quantization operators are also implemented in this module.

\textbf{Downstream Target:}
SAMP supports a lot of models in NLP downstream tasks, including classification, multi-label, named entity recognition, text matching and so on. These capabilities and customization are implemented in Target module.

\subsection{SAMP Transformer-based Encoder}
\begin{figure*}
	\centering
	\subfigure[Fully-Quant mode.]{
		\begin{minipage}[b]{1\textwidth}
			\includegraphics[width=1\textwidth]{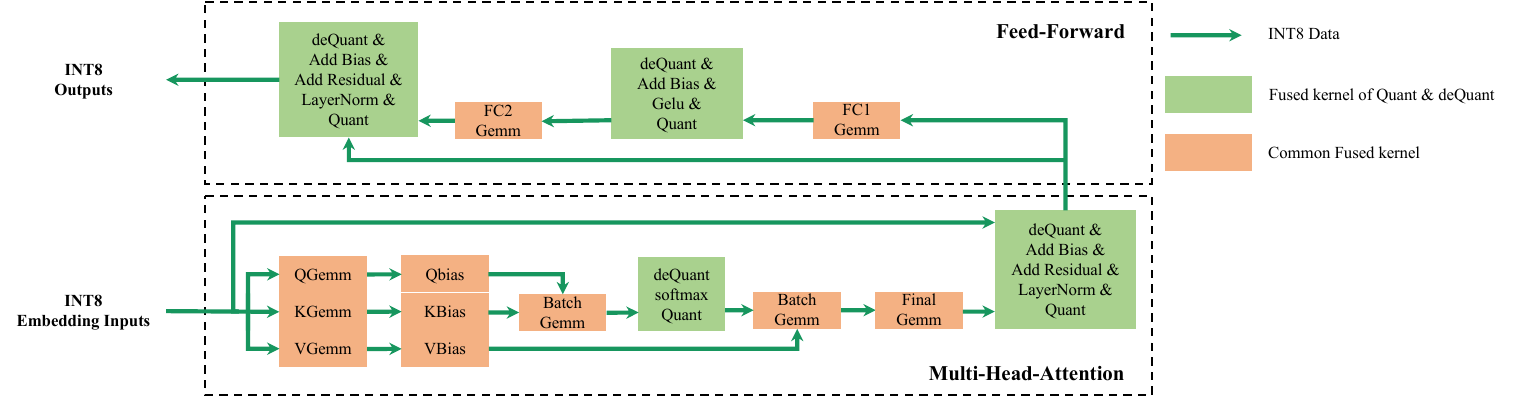}
		\end{minipage}
		\label{fig:samp_fully_int8}
	}
	\\
    \subfigure[Quant-FFN-Only mode.]{
    	\begin{minipage}[b]{1\textwidth}
   		 	\includegraphics[width=1\textwidth]{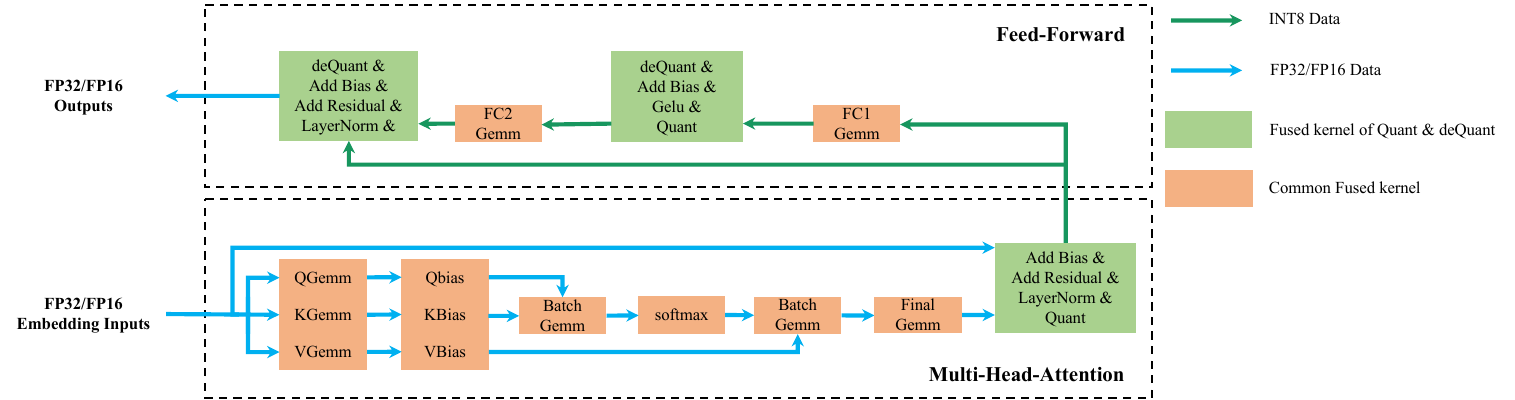}
    	\end{minipage}
		\label{fig:samp_ffn_int8}
    }
	\caption{Two modes of Transformer models in SAMP method. Each square above represents a CUDA kernel, including one or more function operations. Arrows indicate dataflow. For Fully-Quant mode, both Multi-Head Attention and Feed-Forward Network are INT8-quantized in Transformer, while in Quant-FFN-Only mode, only Feed-Forward Network is quantized.}
	\label{fig:samp_quant}
\end{figure*}

In order to solve the problem of serious loss of accuracy, which exists in Fully-quantization method of FasterTransformer, SAMP divides the GEMMs in Transformer Encoder into two categories by multi-head attention (MHA) and feed-forward network (FFN) to generate two mixed-precision modes: Fully-Quant and Quant-FFN-Only. Fully-Quant means GEMMs in MHA and FFN are both quantized. Quant-FFN-Only means GEMMs only in FFN are quantized, and MHA reserves FP32/FP16 accuracy. Figure~\ref{fig:samp_quant} illustrates the kernel details of the two mixed-precision modes in SAMP. For a Transformer-based model, the encoder module is usually composed of lots of Transformer layers. Assuming that the number of layers is L, there are 2L combinations of mixed-precision.


\textbf{Fully-Quant mode} shown in Figure~\ref{fig:samp_fully_int8} quantizes the FP32/FP16 inputs of the Encoder in Embedding module, so the data bit width between Embedding and Encoder is INT8 directly, which reduces the cost of separate quantization kernel call. In addition, we also make a big kernel fusion with Quant/deQuant operations, such as AddResidual, AddBias, and LayerNorm, so that in the whole forward calculation in Encoder, the data transmission between kernels is always 8-bit integer (all green arrows). This fusion reduces the bit width of memory I/O and the number of kernels, making the speed of SAMP INT8-quantization exceed FasterTransformer 5\% $\sim$ 10\%, as shown in Section~\ref{sec:speed_test}.

\textbf{Quant-FFN-Only mode} shown in Figure~\ref{fig:samp_ffn_int8} only quantized the GEMM operations in FFN. As stated above, we reserve the FP32/FP16 GEMM algorithms in MHA, and quantize the floating-point result after LayerNorm operation at the end of MHA. The INT8 GEMM algorithm in FFN is the same as that in Fully-Quant mode, and the only difference is that quantization is not used in the last big kernel to support floating-point outputs.



\begin{algorithm}[t]
    \caption{ Accuracy-Decay-Aware allocator}
    \begin{algorithmic}[1]
    \small
    \REQUIRE $Array$ $A$ , $L$ of Accuracy and Latency, the number of Transformer-layers $N$
    \ENSURE $Lq$, number of quantized Transformer-layers.
        \STATE $dr_{min} \leftarrow $ MAX\_FLOAT
        \STATE $A_{fp16} \leftarrow A_{0}, L_{fp16} \leftarrow L_{0}$
        \FOR{$i=0$ to $N$}
            \IF{$i==0$}
                \STATE $A_{rec} \leftarrow A_{fp16}$
                \STATE $L_{rec} \leftarrow L_{fp16}$
            \ELSE
                \STATE $dr \leftarrow (A_i - A_{rec})/(L_i - L_{rec})$
                \IF{$dr<0$ \textbf{or} $dr<dr_{min}$}
                    \STATE $dr_{min} \leftarrow dr$
                    \STATE $A_{rec} \leftarrow A_i$
                    \STATE $L_{rec} \leftarrow L_i$
                    \STATE $Lq \leftarrow i$
                \ENDIF
            \ENDIF
        \ENDFOR
        \RETURN $Lq$
    \end{algorithmic}
    \label{decay_aware_algo}
\end{algorithm}

We now illustrate how SAMP works effectively. More details of installation and usage are described in Appendix~\ref{sec:appendixA}. For a specific task, SAMP will automatically calculate the accuracy and latency of these mixed-precision combinations of different modes, using Fully-FP16 implementations of SAMP as baseline. Users can input specific latency and accuracy requirements before the calculation. SAMP will find the mixed-precision combination that mostly meets the requirements, and configure the mixed-precision parameters to inference toolkit automatically. When users cannot give clear requirements, SAMP will generate a set of recommended configuration parameters of mixed-precision by the \textbf{Accuracy-Decay-Aware allocation} algorithm. Specifically, in the two modes we proposed above, the speedup increases linearly with the number of quantization layers (each layer of Quant-FFN-Only mode brings 2 $\sim$ 3\% speedup compared with Fully-FP16 in BERT-base binary classification), while the accuracy drops significantly after more layers of quantization. This algorithm will recommend a balance between accuracy and speedup of mixed-precision combination, as shown in Algorithm~\ref{decay_aware_algo}.


\begin{table*}[]
\centering
\resizebox{0.95\textwidth}{!}{
\begin{tabular}{@{}lcccccccc@{}}
\toprule
\multirow{2}{*}{\textbf{PTQ Libraries}} & \multicolumn{2}{c}{\textbf{Quantized Layer}} & \multicolumn{2}{c}{\textbf{AFQMC}} & \multicolumn{2}{c}{\textbf{IFLYTEK}} & \multicolumn{2}{c}{\textbf{TNEWS}} \\ 
                                     & MHA & FFN & Accuracy & Speedup & Accuracy & Speedup & Accuracy & Speedup \\ \cmidrule(r){1-9}
PyTorch-FP16            & 0/12  & 0/12  & 0.7337 & 1.0000 & 0.6048 & 1.0000 & 0.5633 & 1.0000  \\
FasterTransformer-FP16  & 0/12  & 0/12  & 0.7340 & 2.9319 & 0.6052 & 1.6524 & 0.5634 & 3.1351  \\
FasterTransformer-INT8  & 12/12 & 12/12 & 0.5773 & 3.4990 & 0.4540 & 2.4539 & 0.5058 & 3.5551 \\
\textbf{SAMP-FP16}      & 0/12  & 0/12  & 0.7338 & 3.3741 & 0.6056 & 1.4870 & 0.5632 & 3.5022 \\ \cmidrule(r){1-9}
\multirow{6}{*}{\textbf{SAMP-Fully-Quant}}         & 2/12  & 2/12  & 0.6671 & 3.5790 & \underline{0.5572} & \underline{1.5550} & 0.0930 & 3.6790 \\
               & 4/12  & 4/12  & 0.3167 & 3.7689 & 0.2957 & 1.6144 & 0.0856 & 3.9083 \\
               & 6/12  & 6/12  & 0.3188 & 4.0486 & 0.1454 & 1.7305 & \underline{0.0952} & \underline{4.2274} \\
               & 8/12  & 8/12  & 0.6435 & 4.3882 & 0.1493 & 1.8645 & 0.0851 & 4.5985 \\
               & 10/12  & 10/12 & \underline{0.6874} & \underline{4.7751} & 0.1149 & 2.0162 & 0.0900 & 4.9869 \\
               & 12/12 & 12/12 & 0.4409 & 5.1817 & 0.0150 & 2.1978 & 0.0884 & 5.3271 \\ \cmidrule(r){1-9}
\multirow{6}{*}{\textbf{SAMP-Quant-FFN-Only}}          & 0/12  & 2/12  & 0.7340 & 3.4799 & 0.6007 & 1.5073 & 0.5654 & 3.6659 \\
               & 0/12  & 4/12  & 0.7318 & 3.6162 & 0.5932 & 1.5532 & 0.5640 & 3.7465 \\
               & 0/12  & 6/12  & 0.7088 & 3.7725 & 0.5840 & 1.6269 & \underline{0.5610} & \underline{3.9527} \\
               & 0/12  & 8/12  & \underline{0.6872} & \underline{4.0059} & 0.5786 & 1.7095 & 0.5523 & 4.1440 \\
               & 0/12  & 10/12  & 0.5588 & 4.2262 & 0.5663 & 1.7863 & 0.5208 & 4.3917 \\
               & 0/12  & 12/12  & 0.5279 & 4.4574 & \underline{0.5641} & \underline{1.8821} & 0.5077 & 4.6195 \\ \bottomrule
\end{tabular}
}
\caption{\label{citation-guide}
SAMP test for Fine-tuned BERT-base(L12\_H768) model on CLUE tasks AFQMC, IFLYTEK and TNEWS.
We all use min-max calibrator of pytorch-quantization\textsuperscript{\ref{footnote:pytorch_quant}} to generate scales for INT8-quantization. We only show partial experimental data here due to space constraints. Compared with SAMP-FP16, \underline{Underlined scores} represent a mixed-precision combination recommended by the accuracy-decay-aware allocation method in each mode.
}\label{sec:tradeoff_table}
\end{table*}

\section{Experiments}
\subsection{Experiment Settings}
In this section, we show the experimental results of SAMP from two aspects: SAMP trade-off test on text classification tasks and latency speedup. All the evaluation experiments are conducted on GPU NVIDIA Tesla T4, CUDA 11.0. Moreover, we use INT8-quantization calibration tool pytorch-quantization\footnote{\url{https://github.com/NVIDIA/TensorRT/tree/main/tools/pytorch-quantization}\label{footnote:pytorch_quant}} of NVIDIA TensorRT , which provides four calibration methods for post-training quantization (PTQ). Users can choose an appropriate calibration method to generate scale values, which convert model weights from floating-point to fixed-point, for mixed-precision calculations.

First, we test three groups of experiments for SAMP trade-off (between accuracy and latency speedup) in text classification tasks AFQMC(Ant Financial Question Matching Corpus), IFLYTEK(Long Text classification) and TNEWS(Short Text Classification for News) in Chinese Language Understanding Evaluation Benchmark \citep{xu2020clue}. We use BERT-base (12-Layer, HiddenSize-768) released by Google \citep{devlin2018bert} as the pre-trained model, and train the FP32 baseline models by the paradigm of "Pre-training and Fine-tuning" in each task. And we also use TencentPretrain \citep{zhao2022tencentpretrain} as training toolkit. Finally, SAMP self-adaptively obtains the best trade-off between accuracy and latency speedup of mixed-precision on the Dev set.

Secondly, we also test SAMP latency speedup separately, choosing the popular PyTorch and the latest version of FasterTransformer for comparison. Due to the difference of Tokenizer(shown in Table~\ref{sec:toolkit_table}) and programming languages in Target modules (SAMP's targets are developed by C++ programming language, and FasterTransformer uses Python targets \citep{fang2021turbotransformers}), we only make speedup comparison with Encoder.

\subsection{Text Classication on CLUE}
\label{sec:tradeoff_test}
Table~\ref{sec:tradeoff_table} shows the changes of accuracy and speedup with the increase of the number of quantized Transformer-layers in two modes, Fully-Quant and Quant-FFN-Only. The upper bound of speedup is All-layers Fully-Quant and lower bound is Fully-FP16. We choose PyTorch-FP16 implementation as baseline for speedup comparison.
In each mode, with the increase of the number of quantized Transformer-layers, the speedup of three tasks increased steadily, while accuracy decreases faster and faster.

SAMP has three modes: SAMP-FP16, SAMP-Fully-Quant and SAMP-Quant-FFN-Only. It automatically recommends the appropriate mixed-precision combination (underlined scores in Table~\ref{sec:tradeoff_table}) for each task by using the accuracy-decay-aware allocation method. For example, compared with Fully-FP16, in SAMP Quant-FFN-Only mode, the AFQMC task achieves a speedup of 18.7\% (4.0059 vs. 3.3741) through 8-layer FFN quantization, and the accuracy decreases by only 4.7\% (0.6872 vs. 0.7338) . IFLYTEK task achieves a speedup of 26.6\% and accuracy of it decreases by only 4.15\%. In TNEWS task, the accuracy decreases slightly by only 0.22\% in 6-layer FFN quantization, and achieves a speedup of 12.9\%. These recommended results of SAMP have significantly higher accuracy than All-layers Fully-quantization in FasterTransformer, and even most of them have achieved better speedups. Finally, the balance idea between accuracy and latency of SAMP is proved to be effective significantly.

We also find an interesting phenomenon that accuracy decreases heavily in Fully-Quant mode compared with Quant-FFN-Only. The main reason for the severe accuracy loss of quantizing MHA is caused by quantizing the output of Softmax in MHA. For general neural network layers, the distribution of positive and negative outputs are almost balanced, so that the precision range of 8-bit fixed-point ($-2^7$ to $2^7-1$) can be fully used. But the output value of Softmax is between 0 and 1, so the part of -128 to 0 is unused (default in symmetric quantization, refer to Appendix~\ref{sec:appendixB}). Experimental results shows most Softmax output quantized values are distributed between 0 and 64, rather than -128 to 127. The accuracy loss of quantizing Softmax output accumulates when Transformer layer gets deeper, resulting in the overall severe accuracy loss of SAMP Fully-Quant and FasterTransformer. So, \textbf{Quant-FFN-Only is the preferred mode} recommended by SAMP.

\subsection{Speedup of SAMP}
\label{sec:speed_test}
\begin{figure}[ht]
	\centering
	\subfigure[Speedup in Fully-FP32.]{
		\begin{minipage}[b]{0.45\textwidth}
			\includegraphics[width=1\textwidth]{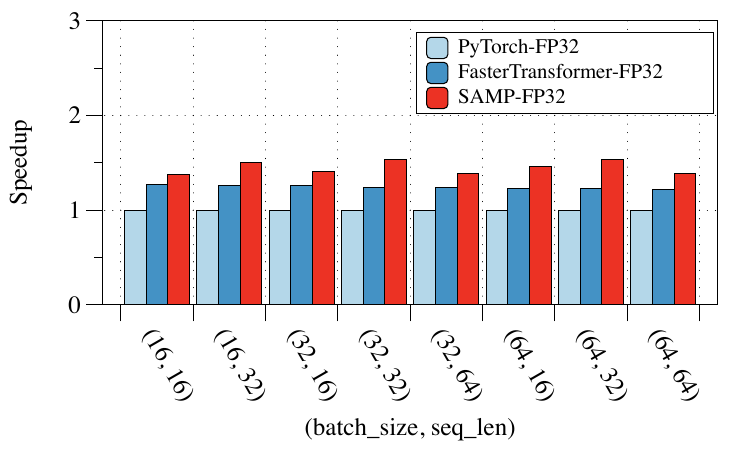}
		\end{minipage}
		\label{fig:samp_speedup_fp32}
	}
    \subfigure[Speedup in Fully-FP16.]{
    	\begin{minipage}[b]{0.45\textwidth}
   		 	\includegraphics[width=1\textwidth]{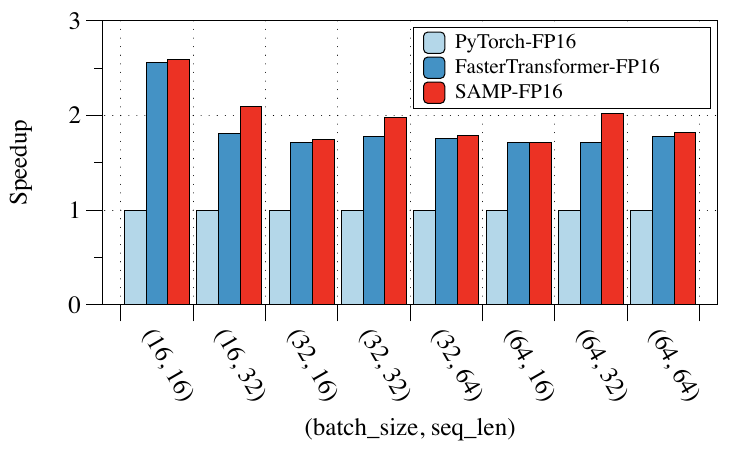}
    	\end{minipage}
		\label{fig:samp_speedup_fp16}
    }
    \subfigure[Speedup in Fully-INT8.]{
        \begin{minipage}[t]{0.45\textwidth}
            \includegraphics[width=1\textwidth]{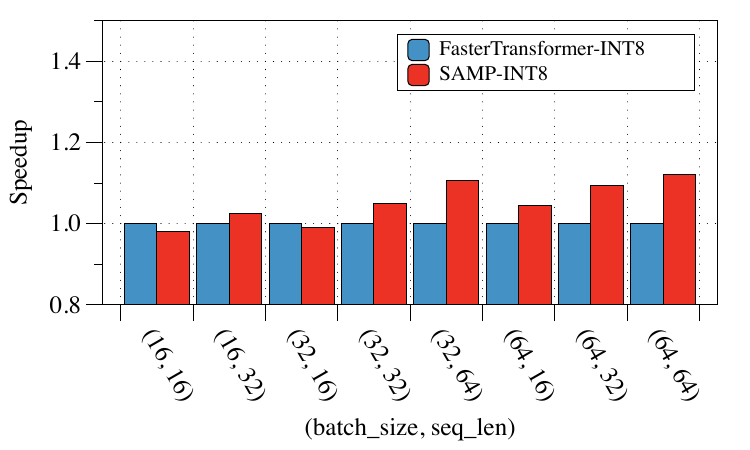}
        \end{minipage}
        \label{fig:samp_speedup_int8}
    }
    \caption{Encoder speedup on GPU Tesla T4 compared with FasterTransformer and PyTorch.}
    \label{fig:samp_speedup}
\end{figure}

For our kernel-fusion improvements, we also make a latency benchmark comparison for fully floating-point and fixed-point, including Fully-FP32, Fully-FP16 and Fully-INT8. As shown in Figure~\ref{fig:samp_speedup}, SAMP floating-point Encoder achieves higher speedups than PyTorch and FasterTransformer in common batch size and length of sequence respectively. These histogram tables show that SAMP-FP32 achieves up to 1.5x speedup compared with PyTorch and 1.1x compared with FasterTransformer, and SAMP-FP16 achieves up to 2x speedup compared with PyTorch and 1.15x compared with FasterTransformer-FP16. Meanwhile, SAMP-Fully-INT8 achieves up to 1.1x speedup compared with FasterTransformer-INT8 in common application scenarios. These comparisons demonstrate that SAMP has been optimized as a new inference tool with faster floating-point and fixed-point computations for Transformer-based Encoder.

\section{Conclusion}
In this paper, we introduce a new inference toolkit SAMP for NLP models. The main contribution of SAMP is to solve the problem of serious performance loss of the existing quantization inference tools in the industrial application of text understanding.
And it also pioneers the application of quantization inference to various downstream tasks
through a wide variety of task-type coverage.
SAMP is light-weight, flexible, and user-friendly. At present, it has been widely used in our business, which greatly saves the deployment cost of industrial applications.

In the future work, we will focus on optimizing the quantization effect of GEMMs in MHA, and explore fixed-point acceleration methods with lower bit width than 8-bit integer, and introduce SAMP to more models.

\nocite{fang2021turbotransformers,wang2021lightseq,vaswani2017attention,devlin2018bert,yang2019xlnet,raffel2020exploring,kim2020fastformers,jacob2018quantization,balzer1991weight,marchesi1993fast,tang1993multilayer,guo2018survey,vanholder2016efficient,nagel2021white,paszke2019pytorch,xu2020clue,abadi2016tensorflow}

\section*{Limitations}

We propose a high-performance quantization inference toolkit SAMP, but it inevitably contains some limitations as:
\begin{itemize}
    \item SAMP is an end-to-end inference toolkit implemented by C++ programming language. Compared with most toolkits of Python programming language, the flexibility of it is limited, and users are required to have some basic knowledge or experience in C++ language development. Based on that, we have provided a lot of convenient Python APIs for ordinary users.
    \item In different series of GPU architectures, the library files of SAMP need to be re-compiled. Users familiar with Nvidia architectures of Data Center know that the construction and compilation of these basic environments are essential.
\end{itemize}

\bibliography{anthology,custom}

\begin{thebibliography}{18}
\expandafter\ifx\csname natexlab\endcsname\relax\def\natexlab#1{#1}\fi

\bibitem[{Abadi et~al.(2016)Abadi, Barham, Chen, Chen, Davis, Dean, Devin,
  Ghemawat, Irving, Isard et~al.}]{abadi2016tensorflow}
Mart{\'\i}n Abadi, Paul Barham, Jianmin Chen, Zhifeng Chen, Andy Davis, Jeffrey
  Dean, Matthieu Devin, Sanjay Ghemawat, Geoffrey Irving, Michael Isard, et~al.
  2016.
\newblock \href
  {https://www.usenix.org/system/files/conference/osdi16/osdi16-abadi.pdf}
  {Tensorflow: a system for large-scale machine learning.}
\newblock In \emph{Osdi}, volume~16, pages 265--283. Savannah, GA, USA.

\bibitem[{Balzer et~al.(1991)Balzer, Takahashi, Ohta, and
  Kyuma}]{balzer1991weight}
Wolfgang Balzer, Masanobu Takahashi, Jun Ohta, and Kazuo Kyuma. 1991.
\newblock \href
  {https://www.sciencedirect.com/science/article/abs/pii/089360809190077I}
  {Weight quantization in boltzmann machines}.
\newblock \emph{Neural Networks}, 4(3):405--409.

\bibitem[{Devlin et~al.(2018)Devlin, Chang, Lee, and
  Toutanova}]{devlin2018bert}
Jacob Devlin, Ming-Wei Chang, Kenton Lee, and Kristina Toutanova. 2018.
\newblock \href {https://arxiv.org/pdf/1810.04805.pdf} {Bert: Pre-training of
  deep bidirectional transformers for language understanding}.
\newblock \emph{arXiv preprint arXiv:1810.04805}.

\bibitem[{Fang et~al.(2021)Fang, Yu, Zhao, and
  Zhou}]{fang2021turbotransformers}
Jiarui Fang, Yang Yu, Chengduo Zhao, and Jie Zhou. 2021.
\newblock \href {https://dl.acm.org/doi/pdf/10.1145/3437801.3441578}
  {Turbotransformers: an efficient gpu serving system for transformer models}.
\newblock In \emph{Proceedings of the 26th ACM SIGPLAN Symposium on Principles
  and Practice of Parallel Programming}, pages 389--402.

\bibitem[{Guo(2018)}]{guo2018survey}
Yunhui Guo. 2018.
\newblock \href {https://arxiv.org/pdf/1808.04752.pdf} {A survey on methods and
  theories of quantized neural networks}.
\newblock \emph{arXiv preprint arXiv:1808.04752}.

\bibitem[{Jacob et~al.(2018)Jacob, Kligys, Chen, Zhu, Tang, Howard, Adam, and
  Kalenichenko}]{jacob2018quantization}
Benoit Jacob, Skirmantas Kligys, Bo~Chen, Menglong Zhu, Matthew Tang, Andrew
  Howard, Hartwig Adam, and Dmitry Kalenichenko. 2018.
\newblock \href
  {https://openaccess.thecvf.com/content_cvpr_2018/papers/Jacob_Quantization_and_Training_CVPR_2018_paper.pdf}
  {Quantization and training of neural networks for efficient
  integer-arithmetic-only inference}.
\newblock In \emph{Proceedings of the IEEE conference on computer vision and
  pattern recognition}, pages 2704--2713.

\bibitem[{Kim and Awadalla(2020)}]{kim2020fastformers}
Young~Jin Kim and Hany~Hassan Awadalla. 2020.
\newblock \href {https://arxiv.org/pdf/2010.13382.pdf} {Fastformers: Highly
  efficient transformer models for natural language understanding}.
\newblock \emph{arXiv preprint arXiv:2010.13382}.

\bibitem[{Marchesi et~al.(1993)Marchesi, Orlandi, Piazza, and
  Uncini}]{marchesi1993fast}
Michele Marchesi, Gianni Orlandi, Francesco Piazza, and Aurelio Uncini. 1993.
\newblock \href
  {http://www.uncini.com/research_activity/pdf/034_ieee_tr_nn4_93.pdf} {Fast
  neural networks without multipliers}.
\newblock \emph{IEEE transactions on Neural Networks}, 4(1):53--62.

\bibitem[{Nagel et~al.(2021)Nagel, Fournarakis, Amjad, Bondarenko, van Baalen,
  and Blankevoort}]{nagel2021white}
Markus Nagel, Marios Fournarakis, Rana~Ali Amjad, Yelysei Bondarenko, Mart van
  Baalen, and Tijmen Blankevoort. 2021.
\newblock \href {https://arxiv.org/pdf/2106.08295.pdf} {A white paper on neural
  network quantization}.
\newblock \emph{arXiv preprint arXiv:2106.08295}.

\bibitem[{Paszke et~al.(2019)Paszke, Gross, Massa, Lerer, Bradbury, Chanan,
  Killeen, Lin, Gimelshein, Antiga et~al.}]{paszke2019pytorch}
Adam Paszke, Sam Gross, Francisco Massa, Adam Lerer, James Bradbury, Gregory
  Chanan, Trevor Killeen, Zeming Lin, Natalia Gimelshein, Luca Antiga, et~al.
  2019.
\newblock \href
  {https://proceedings.neurips.cc/paper/2019/file/bdbca288fee7f92f2bfa9f7012727740-Paper.pdf}
  {Pytorch: An imperative style, high-performance deep learning library}.
\newblock \emph{Advances in neural information processing systems}, 32.

\bibitem[{Raffel et~al.(2020)Raffel, Shazeer, Roberts, Lee, Narang, Matena,
  Zhou, Li, Liu et~al.}]{raffel2020exploring}
Colin Raffel, Noam Shazeer, Adam Roberts, Katherine Lee, Sharan Narang, Michael
  Matena, Yanqi Zhou, Wei Li, Peter~J Liu, et~al. 2020.
\newblock \href
  {https://www.jmlr.org/papers/volume21/20-074/20-074.pdf?ref=https://githubhelp.com}
  {Exploring the limits of transfer learning with a unified text-to-text
  transformer.}
\newblock \emph{J. Mach. Learn. Res.}, 21(140):1--67.

\bibitem[{Tang and Kwan(1993)}]{tang1993multilayer}
Chuan~Zhang Tang and Hon~Keung Kwan. 1993.
\newblock \href {https://ieeexplore.ieee.org/abstract/document/229903}
  {Multilayer feedforward neural networks with single powers-of-two weights}.
\newblock \emph{IEEE Transactions on Signal Processing}, 41(8):2724--2727.

\bibitem[{Vanholder(2016)}]{vanholder2016efficient}
Han Vanholder. 2016.
\newblock Efficient inference with tensorrt.
\newblock In \emph{GPU Technology Conference}, volume~1, page~2.

\bibitem[{Vaswani et~al.(2017)Vaswani, Shazeer, Parmar, Uszkoreit, Jones,
  Gomez, Kaiser, and Polosukhin}]{vaswani2017attention}
Ashish Vaswani, Noam Shazeer, Niki Parmar, Jakob Uszkoreit, Llion Jones,
  Aidan~N Gomez, {\L}ukasz Kaiser, and Illia Polosukhin. 2017.
\newblock \href
  {https://proceedings.neurips.cc/paper/2017/file/3f5ee243547dee91fbd053c1c4a845aa-Paper.pdf}
  {Attention is all you need}.
\newblock \emph{Advances in neural information processing systems}, 30.

\bibitem[{Wang et~al.(2021)Wang, Xiong, Wei, Wang, and Li}]{wang2021lightseq}
Xiaohui Wang, Ying Xiong, Yang Wei, Mingxuan Wang, and Lei Li. 2021.
\newblock \href {https://aclanthology.org/2021.naacl-industry.15.pdf}
  {{L}ight{S}eq: A high performance inference library for transformers}.
\newblock In \emph{Proceedings of the 2021 Conference of the North American
  Chapter of the Association for Computational Linguistics: Human Language
  Technologies: Industry Papers (NAACL-HLT)}, pages 113--120. Association for
  Computational Linguistics.

\bibitem[{Xu et~al.(2020)Xu, Hu, Zhang, Li, Cao, Li, Xu, Sun, Yu, Yu
  et~al.}]{xu2020clue}
Liang Xu, Hai Hu, Xuanwei Zhang, Lu~Li, Chenjie Cao, Yudong Li, Yechen Xu, Kai
  Sun, Dian Yu, Cong Yu, et~al. 2020.
\newblock \href {https://arxiv.org/pdf/2004.05986.pdf} {Clue: A chinese
  language understanding evaluation benchmark}.
\newblock \emph{arXiv preprint arXiv:2004.05986}.

\bibitem[{Yang et~al.(2019)Yang, Dai, Yang, Carbonell, Salakhutdinov, and
  Le}]{yang2019xlnet}
Zhilin Yang, Zihang Dai, Yiming Yang, Jaime Carbonell, Russ~R Salakhutdinov,
  and Quoc~V Le. 2019.
\newblock \href
  {https://proceedings.neurips.cc/paper/2019/file/dc6a7e655d7e5840e66733e9ee67cc69-Paper.pdf}
  {Xlnet: Generalized autoregressive pretraining for language understanding}.
\newblock \emph{Advances in neural information processing systems}, 32.

\bibitem[{Zhao et~al.(2022)Zhao, Li, Hou, Zhao, Tian, Liu, Chen, Sun, Liu, Mao
  et~al.}]{zhao2022tencentpretrain}
Zhe Zhao, Yudong Li, Cheng Hou, Jing Zhao, Rong Tian, Weijie Liu, Yiren Chen,
  Ningyuan Sun, Haoyan Liu, Weiquan Mao, et~al. 2022.
\newblock \href {https://arxiv.org/pdf/2212.06385.pdf} {Tencentpretrain: A
  scalable and flexible toolkit for pre-training models of different
  modalities}.
\newblock \emph{arXiv preprint arXiv:2212.06385}.

\end{thebibliography}
\bibliographystyle{acl_natbib}

\appendix
\section{Installation and Usage}
SAMP is a high performance inference toolkit with less dependencies and high compatibility. 
Pre-request of SAMP only includes CMake >= 3.13, GCC >= 8.3 and CUDA >= 11.0. 
To run calibration, SAMP provides calibration tools depending on PyTorch >= 1.7.0 and NVIDIA pytorch-quantization. 
To install SAMP, users only need to sepcify the GPU compute capability, make a 'build' directory and run CMake and Make.
The executable files will be generated under 'build/bin' directory.

SAMP provide a user-friendly end-to-end inference usage.
First, we can use calibration tools to load the pre-trained language model weights of HuggingFace style and run the calibration process, and dump the weights into a format required by CUDA.
Second, to run self-adaptive mix-precision, SAMP provides some scripts that calibrate the model of different quantization layer settings and recommands the high accuracy and low latency ones.
Under normal conditions, with the number of quantized Transformer-layers gets higher, the model tends to have lower latency but suffers higher accuracy loss. 
Users can set required highest time cost threshold or lowest accuracy threshold. 
If highest time cost threshold is set, SAMP will recommand the setting with the highest accuracy whose time cost is lower than the threshold. 
If the lowest accuracy threshold is set, SAMP will recommand the setting with the lowest time cost whose accuracy is higher than the threshold. 
If neither is set, SAMP will recommand top-5 appropriate settings based on the ratio of speedup / accuracy-loss.
\label{sec:appendixA}

\section{Loss in quantization after Softmax}
\label{sec:appendixB}
Softmax operation is computed during self-attention in transformer models. The output of such operation is quite different from output of other layers.
Common quantization methods usually multiply the floating point numbers by a scale and round them into an integer. 
The rounding operation makes some different floating point numbers rounded into a same integer, which causes the accuracy loss.
To reduce such loss, calibration methods try to find the appropriate scales that makes the quantized integer in -128 to 127 as well-distributed as possible.
The output values of Softmax operation are between 0 and 1. On the premise of symmetric quantization, the part of -128 to 0 is unused. 
In addition, in the output matrix of Softmax, the sum of the elements in the same row is 1, so the attention-softmax output matrix in short sequences tend to have larger element values.
Since the scale in the same layer is pre-computed in calibration process and is fixed in inference process, it is unable to reconcile the distribution of Softmax output in short sequences and long sequences.
We counted the distribution of Attention-Softmax outputs, and take the distribution of MHA output as a comparison. 
Figure-4 shows that the distribution of quantized Attention-Softmax outputs are squeezed in a narrow space of 0-64, while the quantized output of MHA is distributed in -128 to 127.
The accuracy loss of quantized Attention-Softmax outputs accumulate when Transformer layers get deeper, resulting in the overall severe accuracy loss of Fully-quantized SAMP and FasterTransformer.

In the quantized Attention-Softmax outputs, the unused number of INT8-integer is 67.58\%, while in the quantized outputs of MHA, the number is only 4.30\%.

\begin{figure}[ht]
	\centering
	\subfigure[Distribution of quantized MHA outputs]{
		\begin{minipage}[b]{0.45\textwidth}
			\includegraphics[width=1\textwidth]{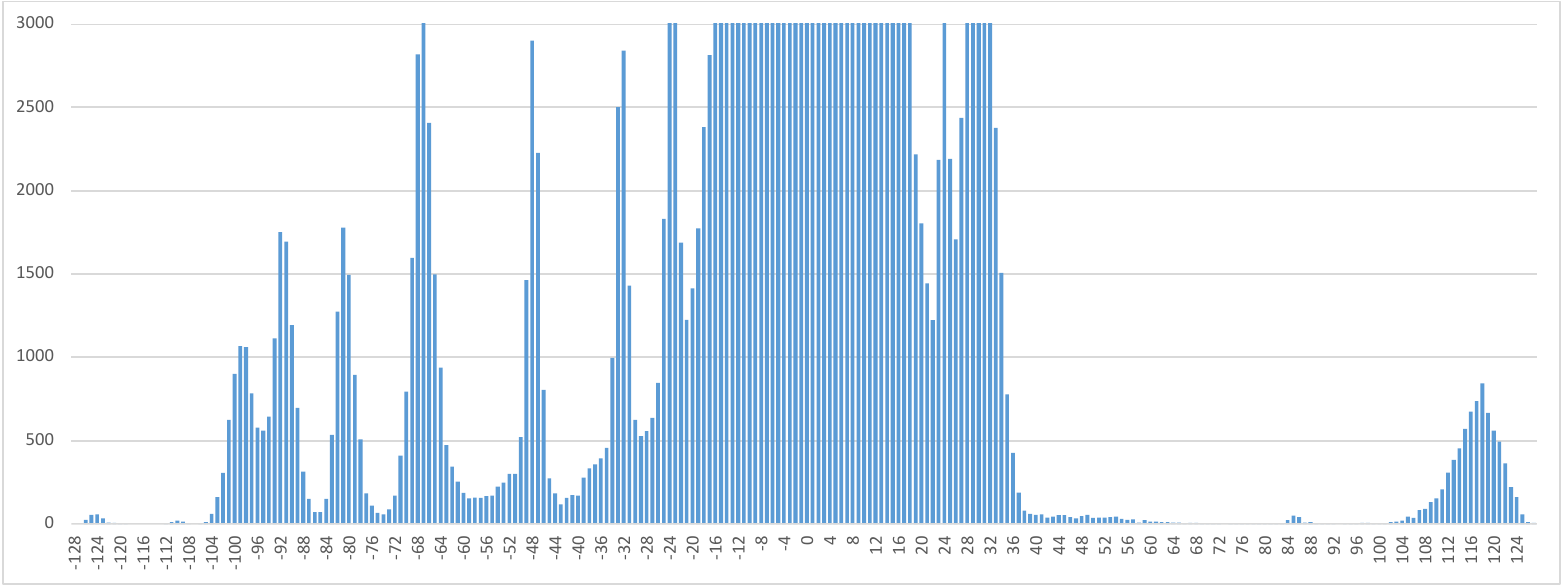}
		\end{minipage}
		\label{fig:mha_output}
	}
    \subfigure[Distribution of quantized Attention-Softmax outputs.]{
    	\begin{minipage}[b]{0.45\textwidth}
   		 	\includegraphics[width=1\textwidth]{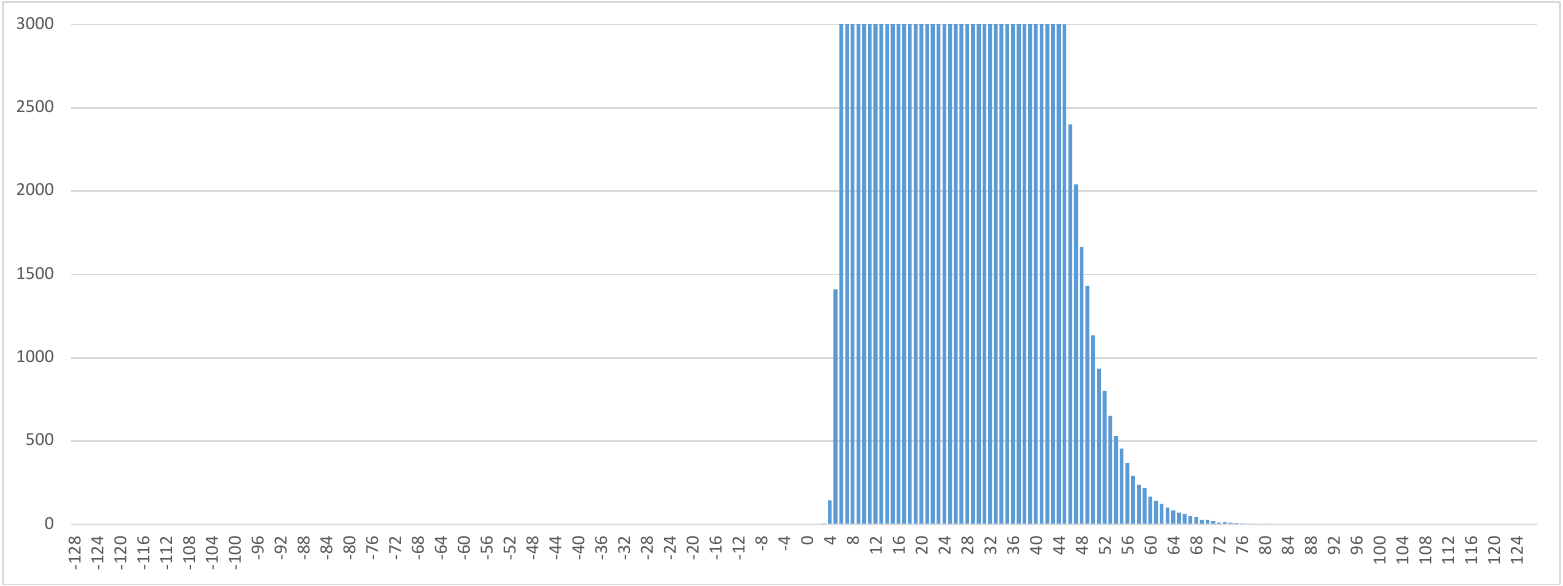}
    	\end{minipage}
		\label{fig:mha_softmax_output}
    }
    \caption{Distribution of quantized MHA output and quantized Attention-Softmax outputs. We count the distribution on 64 sequences of TNEWS classification data. X-axis represents the quantized INT-8 integer values, and Y-axis represents the number of output elements that use the corresponding INT-8 integer values.}
    \label{fig:quant_value_diff}
\end{figure}

\end{document}